\def\year{2018}\relax
\documentclass[letterpaper]{article} 
\usepackage{aaai18}  
\usepackage{times}  
\usepackage{helvet}  
\usepackage{courier}  
\usepackage{url}  
\usepackage{graphicx}  
\usepackage{pdfpages}
\usepackage{amsmath}
\usepackage{amsthm}
\usepackage{amssymb}
\usepackage{comment}
\usepackage{algorithm,algpseudocode}
\theoremstyle{definition}
\newtheorem{defn}{Definition}
\newtheorem{rem}{Remark}

\title{Approximate Vanishing Ideal via Data Knotting}
\author{Hiroshi Kera \and Yoshihiko Hasegawa\\
Department of Information and Communication Engineering, \\
Graduate School of Information Science and Technology, \\
The University of Tokyo, Tokyo, Japan\\
kera@biom.t.u-tokyo.ac.jp
}

\begin{document}

\maketitle
\begin{abstract}
The vanishing ideal is a set of polynomials that takes zero value on the given
data points. Originally proposed in computer algebra, the vanishing ideal has been recently exploited for extracting the nonlinear structures of data in many applications. To avoid overfitting to noisy data,
the polynomials are often designed to approximately rather than exactly equal
zero on the designated data. Although such approximations empirically demonstrate high performance, the sound algebraic structure of the vanishing
ideal is lost. The present paper proposes a vanishing ideal that is tolerant to noisy data and also pursued to have a better algebraic structure. As a new problem, we simultaneously find a set
of polynomials and data points for which the polynomials approximately
vanish on the input data points, and almost exactly vanish on
the discovered data points. In experimental classification tests, our method
discovered much fewer and lower-degree polynomials than an existing state-of-the-art method. Consequently, our method accelerated the runtime of the classification tasks without degrading the classification accuracy.
\end{abstract}

\section{Introduction}

Bridging computer algebra and various applications such as machine learning, computer vision, and systems biology has been attracting interest over the past decade~\cite{torrente2009application,laubenbacher2009computer,li2011theory,livni2013vanishing,vera2014algebra,gao2016nonlinear}.
Borrowed from computer algebra, the vanishing ideal concept has
recently provided new perspectives and has proven effective
in various fields. Especially, vanishing ideal based approaches can discover the nonlinear structure of given data~\cite{laubenbacher2004computational,livni2013vanishing,hou2016discriminative,kera2016noise,kera2016vanishing}. The vanishing ideal is defined as a set of polynomials that always take zero value, \emph{i.e.}, vanishes, on a set of given data points:
\begin{align}
\mathcal{I}(X) & =\left\{ g\in\mathcal{P}\mid g(\mathbf{x})=0,\forall\mathbf{x}\in X\right\},\label{eq:def-vanishing-ideal}
\end{align}
where $X$ is a set of $d$-dimensional data points, and $\mathcal{P}$
is a set of $d$-variate polynomials. A vanishing ideal can be spanned by a finite set of vanishing polynomials~\cite{cox1992ideals}. This basis of the vanishing ideal can be viewed as a system to be satisfied by the input data points; as such, it holds the nonlinear structure of the data. Exploiting these properties, \citeauthor{livni2013vanishing}~(\citeyear{livni2013vanishing}) proposed Vanishing Component Analysis (VCA), which extracts the compact nonlinear
features of the data to be classified, and enables training by a simple
linear classifier. The vanishing ideal has also identified the underlying dynamics
in limited observations~\cite{torrente2009application,robbiano2010approximate,kera2016noise}. In these applications, monomials consisting of models are inferred from the vanishing ideal of the observations.
\begin{figure}
\includegraphics[width=\columnwidth]{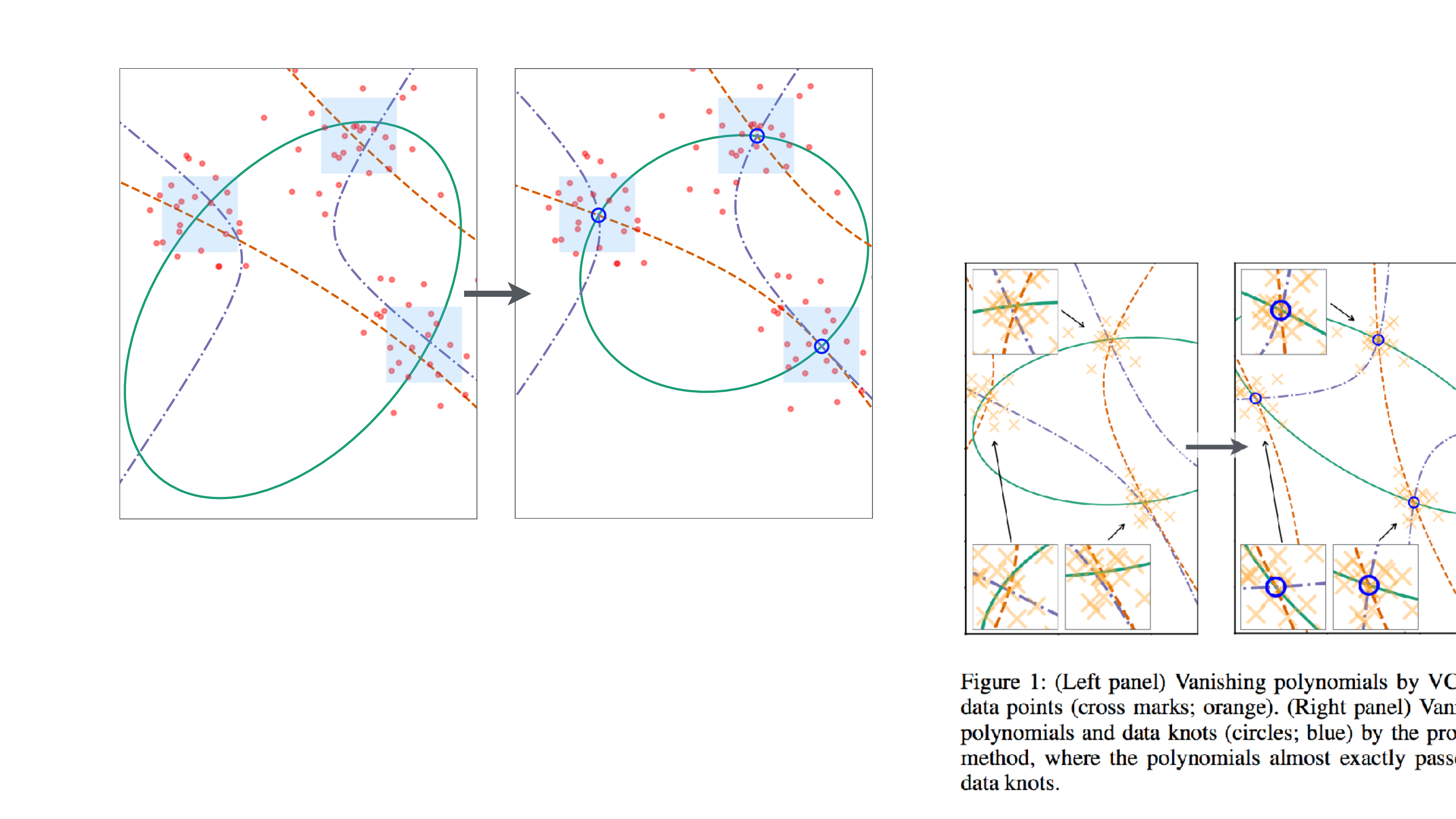}\caption{(Left panel)
Polynomials that approximately vanish on the data points (red dots) in VCA. (Right panel) Vanishing polynomials and data knots (blue circles) obtained by the proposed method, which almost exactly pass the data knots.\label{fig:teaser}}
\end{figure}

As the available data in many applications are exposed to noise, it has been common to compute polynomials that approximately rather than exactly vanish on the data to avoid the overfitting problem~\cite{heldt2009approximate,fassino2010almost,livni2013vanishing,limbeck2014computation}.
However, this approximation destroys the sound algebraic structure of
the vanishing ideal. For instance, as shown in the left panel of Fig.~\ref{fig:teaser},
a set of approximately vanishing polynomials is no longer an algebraic system because it possesses no common roots. In other words, there is a tradeoff between preserving the algebraic soundness of the vanishing ideal and avoiding overfitting
to noise. Such a tradeoff has not been explicitly considered
in existing work.

In the present paper, we newly deal with this tradeoff by addressing a vanishing ideal that is well tolerant to noisy data while pursuing to preserve a sound algebraic structure. Specifically, we introduce a new task that jointly discovers a
set of polynomials and \textit{summarized} data points (called data
knots) from the input data. As shown in the right panel of Fig.~\ref{fig:teaser}, the polynomials in this task avoid overfitting because they approximately vanish on the original data, and preserve the better algebraic structure of a vanishing ideal than those by VCA because they intersect (and vanish) almost exactly at the data knots.

To our knowledge, we present the first
computation of a vanishing ideal that handles both the given fixed points and the jointly
updated points. As the noise level increases, the vanishing ideal of the given fixed data
performs less accurately because it
requires a coarser approximation. If the approximation remains fine, the computed vanishing polynomials
(especially the higher-degree polynomials) will overfit to noise. To circumvent this problem, the proposed method conjointly computes the
vanishing polynomials and the data knots. Assuming that lower-degree polynomials
are less overfitted to noise and better preserve the data structure, our method generates
the vanishing polynomials from lower degree to higher degree while updating
the data knots at each degree. Consequently, overfitting by the higher-degree polynomials
is avoided. The data knots are updated by a nonlinear regularization,
which can be regarded as a generalization
of the Mahalanobis distance, which is commonly used in the metric learning~\cite{bellet2013survey}.
In a theoretical analysis, we also guarantee that the proposed algorithm
terminates and that in the extreme case, the
generated polynomials exactly (rather than almost) vanish on the data knots.

Experiments confirmed that our methods generate fewer and lower-degree polynomials
than an existing vanishing ideal-based approach. In different classification
tasks, our methods obtained a much more compact nonlinear feature than
VCA, reducing the test runtime. To verify that the obtained data
knots well represent the original data, we trained a $k$-nearest neighbor classifier.
Although the data knots are far fewer than the original points,
the classification accuracy of the nearest neighbor classifier was comparable to that of the baseline methods in most cases. 

\section{Related Work}
Although our work addresses a new problem, several works are
closely related to the present study.

\citeauthor{fassino2010almost}~(\citeyear{fassino2010almost}) proposed a Numerical Buchberger--M\"oller (NBM)
algorithm that computes approximately vanishing polynomials for the
vanishing ideal of input data $X$. NBM
requires that each polynomial exactly vanishes on a set of nearby data points $X^{\varepsilon}$ (called an admissible perturbation) from $X$ up to a hyperparameter $\varepsilon$.
However, because the admissible perturbations can differ among the approximately vanishing polynomials,
NBM does not generally output vanishing polynomials that vanish on the same points. In addition, NBM does not specify the admissible perturbations, but only checks the sufficient condition of the existence of such points.

\citeauthor{Abbott2007}~(\citeyear{Abbott2007}) addressed a new task of \emph{thinning
out} the data points as a preprocessing for computing vanishing ideals
afterward. Similar to clustering, their
approach computes the empirical centroids of the input data. As thinning
out and the vanishing-ideal computation are performed independently, the
empirical centroids do not necessarily result in compact, lower-degree
vanishing polynomials. In contrast, our method conjointly performs the thinning out and vanishing-ideal computation in a single framework. Note that both methods by \citeauthor{Abbott2007} and by us aim to reduce data points by summarizing nearby points, which are different from the clustering tasks that aim to group even distant points according to their categories.


\section{Preliminaries}
\begin{defn}
(Vanishing Ideal) Given a set of $d$-dimensional data points $X$, the vanishing ideal of $X$ is a set of $d$-variate polynomials that take zero values, (\emph{i.e.}, vanish) at all points in $X$. Formally,
the vanishing ideal is defined as Eq.~(\ref{eq:def-vanishing-ideal}).%
\end{defn}

\begin{defn}
(Evaluation vector, Evaluation matrix) Given a set of data points
$X=\left\{ \mathbf{x}_{1},\mathbf{x}_{2},\ldots,\mathbf{x}_{N}\right\} $
with $N$ samples, the evaluation vector of polynomial $f$ is defined as
\begin{align*}
f(X) & =\begin{pmatrix}f(\mathbf{x}_{1}) & f(\mathbf{x}_{2}) & \cdots & f(\mathbf{x}_{N})\end{pmatrix}^{\top}\in\mathbb{R}^{N}.
\end{align*}
Given a set of polynomials $F=\left\{ f_{1},f_{2},...,f_{s}\right\} $,
its evaluation matrix is $F(X)=(f_{1}(X)\ f_{2}(X)\ \cdots\ f_{s}(X))\in\mathbb{R}^{N\times s}$.
\end{defn}

\begin{defn}
($\varepsilon$-vanishing polynomial) Given a set of data points $X$,
a polynomial $g$ is called an $\varepsilon$-vanishing polynomial
on $X$ when the norm of its evaluation vector on $X$ is less than or
equal to $\varepsilon$, \emph{i.e.}, $\|g(X)\|\le\varepsilon$, where $\|\cdot\|$
denotes the Euclidean norm of a vector. Otherwise, $g$ is called an
$\varepsilon$-nonvanishing polynomial.
\end{defn}

In the vanishing ideal concept, polynomials are identified with their evaluation on the points; that is, a polynomial
$f$ is mapped to an $N$-dimensional vector $f(X)$. Two polynomials $f$ and $\tilde{f}$ are considered to be equivalent if they have the same evaluation vector, \emph{i.e.}, $f(X)=\tilde{f}(X)$. When $f(X)=\mathbf{0}$, $f$ is a vanishing polynomial on $X$.

Another important relation between polynomials and their evaluations is that the sum of the evaluation of polynomials equals to the evaluation of the sum of the polynomials. More specifically, the following relation holds for a set of polynomials $G=\left\{ g_{1},...,g_{s}\right\} $ and coefficient vectors $\mathbf{c}\in\mathbb{R}^{s}$: 
\begin{align*}
G(X)\mathbf{c} & =(G\mathbf{c})(X),
\end{align*}
where $G\mathbf{c}=\sum_{i=1}^{s}c_{i}g_{i}$ defines the inner product between a set $G$ and a coefficient
vector $\mathbf{c}$, with
$c_{i}$ being the $i$-th entry of $\mathbf{c}$. Similarly,
$G\mathbf{C}=\left\{ G\mathbf{c}_{1},...,G\mathbf{c}_{k}\right\} $ defines the multiplication of $G$ by a matrix $\mathbf{C}=\left(\mathbf{c}_{1},\cdots,\mathbf{c}_{k}\right)$.
These special inner products will be used hereafter.

\section{Proposed Method}
Given a set of data points $X_{0}$, we seek a set of polynomials
$G$ and data knots $Z$ such that the polynomials in $G$ approximately
vanish on $X_{0}$ and almost exactly vanish on $Z$. Specifically,
\begin{align}
\forall g\in G, & \begin{Vmatrix}g(X_{0})\end{Vmatrix}\le\varepsilon,\begin{Vmatrix}g(Z)\end{Vmatrix}\le\delta,\label{eq:approx-exact-vanishing-polynomials}
\end{align}
where $\mathbf{\varepsilon},\delta$ are hyperparameters of the error
tolerance that requires $g$ to be $\varepsilon$-vanishing on $X_{0}$
and $\delta$-vanishing on $Z$. In our case, $\delta$ is set much smaller than $\varepsilon$.

Like many other methods that construct a vanishing ideal basis, our polynomial
construction starts from a degree-$0$
polynomial (\emph{i.e.}, constant), and increments the polynomial degree in successive data fittings. For each degree $t$, the proposed method
iterates the following two steps:
\begin{enumerate}
\item Compute a set of approximately vanishing polynomials $G_{t}$ and
a set of nonvanishing polynomials $F_{t}$.
\item Update the data knots such that the approximately vanishing polynomials
in $G^{t}=\cup_{i=1}^{t}G_{i}$ more closely vanish.
\end{enumerate}
These steps constitute our technical contributions to the field. In the former step, we newly introduce a vanishing polynomial construction that handles two set of points $X_{0},Z$ with different approximation scales $\varepsilon,\delta$.
In the latter step, we introduce a nonlinear regularization term that preserves the nonlinear structure of data while
updating $Z$.

In the following subsections, we first describe the polynomial construction part (Step~1; Section~\ref{subsec:Non-vanishing-and-vanishing})
and the data-knotting part (Step~2; Section~\ref{subsec:Data-knotting}).
To improve the quality of the vanishing polynomials and data knots, we propose an iterative update framework in Section~\ref{subsec:Exact-Vanish-Pursuit}. Finally, we present our whole algorithm in Section~\ref{subsec:Algorithm}.

\subsection{Vanishing polynomial construction\label{subsec:Non-vanishing-and-vanishing}}

\begin{algorithm}[t]
\caption{FindBasis\label{alg:FindBasis}}
\begin{algorithmic}[1]
\Require{
$
\tilde{C}_t, F^{t-1}, Z, X_0, \varepsilon, \delta$
}
\Ensure{$G_t, F_t$}
\State{\# Compute residual space}
\State{$C_t = \tilde{C}_t - F^{t-1}F^{t-1}(Z)^{\dagger}\tilde{C}_t(Z)$}

\State{}
\State{\# SVD for evaluation matrix on $X_0$}
\State{$C_t(X_0) = \mathbf{U}_0\mathbf{D}_0\mathbf{V}_0^{\top} = \mathbf{U}_0\mathbf{D}_0[\tilde{\mathbf{V}}_0\ \mathbf{V}_0^{\varepsilon}]^{\top}$}
\State{}

\State{\# SVD for evaluation matrix on $Z$}
\State{$C_t(Z)\mathbf{V}_0^{\varepsilon} = \mathbf{U}\mathbf{D}\mathbf{V}^{\top}= \mathbf{U}\mathbf{D}[\tilde{\mathbf{V}}\ \mathbf{V}^{\delta}]^{\top}$}
\State{$C_t(Z)\tilde{\mathbf{V}}_0 = \hat{\mathbf{U}}\mathbf{E}\mathbf{W}^{\top} = \hat{\mathbf{U}}\tilde{\mathbf{E}}[\tilde{\mathbf{W}}\ \mathbf{W}^{\delta}]^{\top}$}
\State{}

\State{$G_t = C_{t}\mathbf{V}_{0}^{\varepsilon}\mathbf{V}^{\delta}$}
\State{$F_t=C_{t}\tilde{\mathbf{V}}_{0}\tilde{\mathbf{W}}\cup C_{t}\mathbf{V}_{0}^{\varepsilon}\tilde{\mathbf{V}}$}
\State{\# Divide each polynomial in $F_t$ by the norm of its evaluation vectors on $Z$ }
\State{\Return $G_t, F_t$}
\end{algorithmic}
\end{algorithm}

Given a set of data points $X_{0}$ and tentative data knots $Z$,
we here construct two polynomial sets $G_{t}$ and
$F_{t}$ of degree $t\ge 1$. $G_{t}$ is the set of degree-$t$
polynomials that are $\varepsilon$-vanishing on $X_{0}$ and $\delta$-vanishing
on $Z$. $F_{t}$ is a set of polynomials that are not $\delta$-vanishing on $Z$. Note that at this moment, we have $G^{t-1}=\cup_{i=0}^{t-1}G_{i}$
and $F^{t-1}=\cup_{i=0}^{t-1}F_{i}$.

To obtain $G_{t}$ and $F_{t}$, we first generate candidate degree-$t$ polynomials $C_{t}$ based on the VCA framework. Multiplying all possible combinations of polynomials between $F_{1}$ and $F_{t-1}$, we construct $\tilde{C}_{t} =\left\{ fg\mid f\in F_{1},g\in F_{t-1}\right\}$. Then, $C_{t}$ is generated as the residual polynomials of $\tilde{C}_{t}$ with respect to $F^{t-1}$ and their evaluation vectors. 
\begin{align*}
C_{t} & =\tilde{C}_{t}-F^{t-1}\left(F^{t-1}(Z)^{\dagger}\tilde{C}_{t}(Z)\right),
\end{align*}
where $A^{\dagger}$ denotes the pseudo-inverse matrix of $A$.
Note that the second term is the inner product between $F^{t-1}$
and $F^{t-1}(Z)^{\dagger}\tilde{C}_{t}(Z)$, and not the evaluation
of $F^{t-1}$ on $F^{t-1}(Z)^{\dagger}\tilde{C}_{t}(Z)$. This step calculates the residual
column space of $\tilde{C}_{t}(Z)$ that is orthogonal to that of
$F^{t-1}(Z)$.
\begin{align*}
C_{t}(Z) & =\tilde{C}_{t}(Z)-F^{t-1}(Z)\left(F^{t-1}(Z)^{\dagger}\tilde{C}_{t}(Z)\right).
\end{align*}
In short, when evaluated on the data knots $Z$, the column spaces of the residual polynomials $C_{t}$ and the above polynomials are orthogonal.

After generating $C_{t}$, degree-$t$ nonvanishing polynomials $F_{t}$
and vanishing polynomials $G_{t}$ are constructed by applying singular value
decomposition~(SVD) to $C_{t}(X_{0})$:
\begin{align*}
C_{t}(X_{0}) & =\mathbf{U}_{0}\mathbf{D}_{0}\mathbf{V}_{0}^{\top},
\end{align*}
where $\mathbf{U}_{0}\in\mathbb{R}^{N\times N}$ and $\mathbf{V}_{0}\in\mathbb{R}^{|C|\times|C|}$
are orthogonal matrices, and $D\in\mathbb{R}^{N\times|C|}$ contains singular values only along its diagonal. Multiplying both sides by $\mathbf{V}_{0}$ and focusing on the $i$-th column, we obtain
\begin{align*}
C_{t}(X_{0})\mathbf{v}_{i} & =\sigma_{i}\mathbf{u}_{i},
\end{align*}
where $\mathbf{v}_{i}$ and $\mathbf{u}_{i}$ are the $i$-th columns of
$\mathbf{V}_{0}$ and $\mathbf{U}_{0}$ respectively, and $\sigma_{i}$ is
the $i$-th diagonal entry of $\mathbf{D}_{0}$. Moreover, when $\sigma_{i}\le\mathbf{\varepsilon}$, we have
\begin{align*}
\|g_{i}(X_{0})\| & =\begin{Vmatrix}(C_{t}\mathbf{v}_{i})(X_{0})\end{Vmatrix}=\begin{Vmatrix}C_{t}(X_{0})\mathbf{v}_{i}\end{Vmatrix}=\|\sigma_{i}\mathbf{u}_{i}\|=\sigma_{i},
\end{align*}
meaning that polynomial $g_{i}:=C_{t}\mathbf{v}_{i}$
is an $\varepsilon$-vanishing polynomial on $X_{0}$.
We can regard $\mathbf{v}_{i}$ as a coefficient vector of the linear combination of
the polynomials in $C_{t}$. Let us denote $\mathbf{V}_{0}=(\tilde{\mathbf{V}}_{0}\ \mathbf{V}_{0}^{\varepsilon})$,
where $\tilde{\mathbf{V}}_{0}$ and $\mathbf{V}_{0}^{\varepsilon}$
are matrices corresponding to the singular values exceeding $\varepsilon$
and not exceeding $\varepsilon$, respectively. For any
unit vector $\mathbf{v}=\mathbf{V}_{0}^{\varepsilon}\mathbf{p}\in\text{span}(\mathbf{V}_{0}^{\varepsilon})$ expressed in terms of a unit vector $\mathbf{p}$, a polynomial $g=C_{t}\mathbf{v}$ satisfies
\begin{align*}
\|g(X_{0})\| & =\|C_{t}(X_{0})\mathbf{v}\|=\|G_{t}(X_{0})\mathbf{p}\|\le \sigma_{\varepsilon_\text{max}}\|\mathbf{p}\| =\sigma_{\varepsilon_\text{max}},
\end{align*}
meaning that $g=C_{t}\mathbf{v}$ is an $\varepsilon$-vanishing polynomial, where $\sigma_{\varepsilon_{\text{max}}}$ is the largest singular value not exceeding $\varepsilon$.

Next, we construct $\delta$-vanishing polynomials
in the above polynomial space. This constraint problem is formulated as
follows:
\begin{align*}
\min_{(\hat{\mathbf{V}}^{\delta})^{\top}\hat{\mathbf{V}}^{\delta}=\mathbf{I}} & \|C_{t}(Z)\hat{\mathbf{V}}^{\delta}\|_{\text{F}}, & \text{s.t. }\text{span}(\hat{\mathbf{V}}^{\delta})\subset\text{span}(\mathbf{V}_{0}^{\varepsilon}),
\end{align*}
where $\|\cdot\|_{\text{F}}$ denotes the Frobenius norm of a matrix.
From the discussion above, it can be reformulated as, 
\begin{align*}
\min_{\mathbf{P}^{\top}\mathbf{P}=\mathbf{I}} & \|C_{t}(Z)\mathbf{V}_{0}^{\varepsilon}\mathbf{P}\|_{\text{F}},
\end{align*}
since the column space of $\mathbf{V}_{0}^{\varepsilon}$ spans the coefficient vectors of $\varepsilon$-vanishing polynomials on $X_0$.
This problem can be simply solved by applying SVD to $C_{t}(Z)\mathbf{V}_{0}^{\varepsilon}$
and selecting the right singular vectors corresponding to the singular
values not exceeding $\delta$. Supposing that SVD yields $C_{t}(Z)\mathbf{V}_{0}^{\varepsilon}=\mathbf{U}\mathbf{D}\mathbf{V}^{\top}$,
we denote $\mathbf{V}=(\tilde{\mathbf{V}}\  \mathbf{V}^{\delta})$,
where $\tilde{\mathbf{V}}$ and $\mathbf{V}^{\delta}$ are matrices
corresponding to the singular values exceeding $\delta$
and not exceeding $\delta$, respectively. The polynomials
$G_{t}$ that are $\varepsilon$-vanishing on $X$ and $\delta$-vanishing
on $Z$ are then obtained as follows:
\begin{align*}
G_{t} & =C_{t}\mathbf{V}_{0}^{\varepsilon}\mathbf{V}^{\delta}.
\end{align*}
$F_{t}$ is constructed similarly, but consists of two
parts.
\begin{align*}
F_{t} & =C_{t}\tilde{\mathbf{V}}_{0}\tilde{\mathbf{W}}\cup C_{t}\mathbf{V}_{0}^{\varepsilon}\tilde{\mathbf{V}},
\end{align*}
where $\tilde{\mathbf{W}}$ is the counterpart of $\tilde{\mathbf{V}}$
for $C_{t}(Z)\tilde{\mathbf{V}}_{0}$. The left-part polynomials are
$\varepsilon$-nonvanishing on $X_{0}$ and $\delta$-nonvanishing
on $Z$, whereas the right-part polynomials are $\varepsilon$-vanishing on
$X_{0}$ but $\delta$-nonvanishing on $Z$. Following the VCA framework, the polynomials in $F_{t}$ are rescaled by their evaluation vectors on $Z$ to maintain numerical stability. The main generating procedures of
$G_{t}$ and $F_{t}$ are summarized in Alg.~\ref{alg:FindBasis}.
For simplicity, we omit the step that constructs $\tilde{C}_{t}$.

\subsection{Data knotting\label{subsec:Data-knotting}}

Given a set of points $X$ and a set of vanishing polynomials $G^{t}$ up to degree $t$, we seek new data points $Z$ for which the polynomials in $G^{t}$ more exactly vanish on $Z$ while preserving the nonlinear structure of $X$. In the present paper, we refer to these $Z$ as data knots and the searching process as data knotting (by analogy to ropes). As described later, we iteratively update the data knots, so we designate the current and updated data knots as $X$ and $Z$, respectively.

First, we intuitively illustrate the concept in a simple case. Let
$\mathbf{X}$ to be a matrix whose $i$-th row corresponds to the
$i$-th point of $X$. Applying SVD, we have
\begin{align}
\mathbf{X} & =\mathbf{UDV}^{\top}=\tilde{\mathbf{U}}\tilde{\mathbf{D}}\tilde{\mathbf{V}}^{\top}+\mathbf{U}_{\varepsilon}\mathbf{D}_{\varepsilon}\mathbf{V}_{\varepsilon}^{\top},\label{eq:linear-data-knotting}
\end{align}
where $\mathbf{U}, \mathbf{V}$ are orthonormal matrices, and $\mathbf{D}$ has the singular values only along its diagonal.
The former and latter terms define the principal and minor variances in the data, respectively, regarding singular values exceeding $\varepsilon$ and the rest. Note that when $\mathbf{X}$ is mean-centralized,
$C_{1}(X)=\mathbf{X}$. We seek new data points
for which the minor variance ideally vanishes to a zero matrix while the principal
variance is preserved. In this linear case, the new points are simply defined as $\mathbf{Z}=\tilde{\mathbf{U}}\tilde{\mathbf{D}}\tilde{\mathbf{V}}^{\top}$,
where $\mathbf{Z}$ for $Z$ is defined identically to $\mathbf{X}$.

However, discovering $Z$ in nonlinear cases is not straightforward. In the degree-$t$ case, Eq.~(\ref{eq:linear-data-knotting}) becomes
\begin{align}
C_{t}(X) & =\tilde{\mathbf{U}}\tilde{\mathbf{D}}\tilde{\mathbf{V}}^{\top}+\mathbf{U}_{\varepsilon}\mathbf{D}_{\varepsilon}\mathbf{V}_{\varepsilon}^{\top}.\label{eq:svd-nonlinear}
\end{align}
As in the linear case, we seek $Z$ such that $C_{t}(Z)=\tilde{\mathbf{U}}\tilde{\mathbf{D}}\tilde{\mathbf{V}}^{\top}$. To this end, we address the following minimization problem:
\begin{align}
\min_{Z}\  & \begin{Vmatrix}C_{t}(Z)-\tilde{\mathbf{U}}\tilde{\mathbf{D}}\tilde{\mathbf{V}}^{\top}\end{Vmatrix}_{\text{F}}.\label{eq:original-problem}
\end{align}
 Multiplying Eq.~(\ref{eq:svd-nonlinear}) by $\mathbf{V}$ and $\mathbf{V}_{\varepsilon}$
respectively, we obtain
\begin{align*}
F_{t}(X) & =F_{t}(Z),\\
G_{t}(Z) & =\mathbf{O}.
\end{align*}
In the derivation, we used $\tilde{\mathbf{V}}^{\top}\tilde{\mathbf{V}}=\mathbf{I}$,
$\mathbf{V}_{\varepsilon}^{\top}\mathbf{V}_{\varepsilon}=\mathbf{I}$,
and $\tilde{\mathbf{V}}^{\top}\mathbf{V}_{\varepsilon}=\mathbf{O}$,
where $\mathbf{I}$ is an identity matrix and $\mathbf{O}$ is a zero
matrix. From these relations, Eq.~(\ref{eq:original-problem}) is reformulated as,
\begin{align}
\min_{Z}\  & \|G_{t}(Z)\|_{\text{F}}+\lambda_{t}\|F_{t}(Z)-F_{t}(X)\|_{\text{F}},\label{eq:optimization-problem}
\end{align}
where $\lambda_{t}$ is a hyperparameter. It can be easily
shown, 
\begin{align*}
C_{t}(Z)=\tilde{\mathbf{U}}\tilde{\mathbf{D}}\tilde{\mathbf{V}}^{\top}\iff F_{t}(X)=F_{t}(Z),G_{t}(Z)=\mathbf{O}. 
\end{align*}
See the supplementary material for the proof. Note that the optimization
problem of Eq.~(\ref{eq:optimization-problem}) can be factorized
into a subproblem on each data point.
\begin{align}
\min_{\mathbf{z}_{i}}\  & \|G_{t}(\mathbf{z}_{i})\|_{\text{F}}+\lambda_{t}\|F_{t}(\mathbf{z}_{i})-F_{t}(\mathbf{x}_{i})\|_{\text{F}},\label{eq:optimization-problem-pointwise}
\end{align}
where $\mathbf{z}_{i}, \mathbf{x}_{i}$ are the $i$-th data point
of $Z,X$, respectively. Considering the polynomials up to
degree $t$, Eq.~(\ref{eq:optimization-problem-pointwise}) becomes
\begin{align}
\min_{\mathbf{z}_{i}}\  & \sum_{k=1}^{t}\|G_{k}(\mathbf{z}_{i})\|_{\text{F}}+\sum_{k=1}^{t}\lambda_{k}\|F_{k}(\mathbf{z}_{i})-F_{k}(\mathbf{x}_{i})\|_{\text{F}}.\label{eq:full-optimization-problem-pointwise}
\end{align}
The first term encourage the polynomials in $G^{t}$ to vanish on
$\mathbf{z}_{i}$, and the second term constrains the $\mathbf{z}_{i}$
to nearby $\mathbf{x}_{i}$ with respect to $F^{t}$. This formulation provides an interesting insight:
\begin{rem}
The second term of Eq.~(\ref{eq:full-optimization-problem-pointwise})
is a regularization term that is equivalent to a nonlinear generalization
of the Mahalanobis distance. 
\end{rem}
The Mahalanobis distance between two data points $\mathbf{x}$ and
$\mathbf{y}$ of a mean-centralized matrix $\mathbf{X}$ is a
generalization of the Euclidean distance, where each variable is
normalized by its standard deviation, \emph{i.e.}, $d(\mathbf{x},\mathbf{y})=\sqrt{(\mathbf{x}-\mathbf{y})^{\top}\mathbf{\Sigma}^{\dagger}(\mathbf{x}-\mathbf{y})}$,
where $\mathbf{\Sigma}=\mathbf{X}^{\top}\mathbf{X}$ is an empirical
covariance matrix. The remark above holds because our $F_{t}$ describes the nonlinear principal variance of the current data knots $X$. For simplicity, we consider only the linear case $t=1$. In this case, $C_{1}(X)$ is a mean-centralized $X$ because it is the residual with respect to $F_{0}=1/\sqrt{|X|}$.
Adopting the notations in lines 8 and 9 of Alg.~\ref{alg:FindBasis},
we let $\mathbf{\tilde{E}}$ and $\tilde{\mathbf{D}}$ be submatrices
of $\mathbf{E}$ and $\mathbf{D}$ corresponding to singular values larger
than $\delta$. The distance between two points $\mathbf{x},\mathbf{y}$ of the mean-centralized $X$ is then calculated as 
\begin{align*}
 & \|F_{1}(\mathbf{x})-F_{1}(\mathbf{y})\|^{2}\\
 & =\begin{Vmatrix}\mathbf{x}^{\top}\begin{pmatrix}\tilde{\mathbf{V}}_{0}\tilde{\mathbf{W}}\tilde{\mathbf{E}}^{\dagger} & \mathbf{V}_{0}^{\varepsilon}\tilde{\mathbf{V}}\tilde{\mathbf{D}}^{\dagger}\end{pmatrix}-\mathbf{y}^{\top}\begin{pmatrix}\tilde{\mathbf{V}}_{0}\tilde{\mathbf{W}}\tilde{\mathbf{E}}^{\dagger} & \mathbf{V}_{0}^{\varepsilon}\tilde{\mathbf{V}}\tilde{\mathbf{D}}^{\dagger}\end{pmatrix}\end{Vmatrix}^{2},\\
 & =(\mathbf{x}-\mathbf{y})^{\top}\Sigma^{-1}(\mathbf{x}-\mathbf{y}),
\end{align*}
where 
\begin{align*}
\Sigma^{-1} & =\tilde{\mathbf{V}}_{0}\tilde{\mathbf{W}}(\tilde{\mathbf{E}}^{\top}\tilde{\mathbf{E}})^{-1}\tilde{\mathbf{W}}^{\top}\tilde{\mathbf{V}}_{0}^{\top}+\mathbf{V}_{0}^{\varepsilon}\tilde{\mathbf{V}}(\tilde{\mathbf{D}}^{\top}\tilde{\mathbf{D}})^{-1}\tilde{\mathbf{V}}^{\top}{\mathbf{V}_{0}^{\varepsilon}}^{\top}.
\end{align*}
 A straightforward calculation shows that $\Sigma$ is the principal variance of the empirical covariance matrix $C_{1}(X)^{\top}C_{1}(X)$. Similarly, in nonlinear case $t>1$, $\|F_{t}(\mathbf{x})-F_{t}(\mathbf{y})\|$ is a Mahalanobis distance with respect to principal variance of $C_{t}(X)^{\top}C_{t}(X)$.
Therefore, the regularization term in Eq.~(\ref{eq:full-optimization-problem-pointwise})
can be regarded as a generalized Mahalanobis distance in nonlinear cases (details are provided in the supplementary material).

To optimize Eq.~(\ref{eq:full-optimization-problem-pointwise}), quasi-Newton method with numerical gradient calculation was adopted in the present paper. We restricted the $F_{k}$ to lower-degree polynomials by assuming that lower-degree structures holds sufficiently good structure of data. Specifically, we took into account the regularization terms up to the degree where the first vanishing polynomial is found. 

\subsection{Exact Vanish Pursuit\label{subsec:Exact-Vanish-Pursuit}}

Our original goal to discover the $\delta$-vanishing
polynomials on the data knots cannot be achieved by simply applying the data
knotting to a fixed set of polynomials. In some cases (e.g., when three polynomials never
intersect at any one instant), there may be no data knots on which the given polynomials sufficiently vanish. To resolve this problem, we introduce an iterative framework that alternatively updates data knots and polynomials~(Alg.~\ref{alg:ExactVanishPursuit}). 

Let us construct the degree-$t$ polynomials $G_{t}$ and
$F_{t}$. At this moment, we have polynomial sets with degree less than $t$, $G^{t-1}$ and $F^{t-1}$, and tentative data knots $Z$.
Introducing $\eta>\delta$, we repeat the following steps.
\begin{enumerate}
\item Fixing $Z$, update $G_{t}$ and $F_{t}$ by Alg.~\ref{alg:FindBasis}. In this step, the polynomials in $G_{t}$ are $\varepsilon$-vanishing on $X_{0}$ and $\eta$-vanishing on $Z$.
\item Fixing $F^{t},G^{t}$, update the data knots $Z$ by solving Eq.~(\ref{eq:full-optimization-problem-pointwise}).
\item Decrease $\eta$.
\end{enumerate}
This iteration terminates before $\eta$ reaches $\delta$ when $G^{t}$ becomes a set of $\delta$-vanishing polynomials, $G_{t}$ becomes an empty set. The parameter $\eta$ approaches to $\delta$ over the iterations, and when $\eta=\delta$ then all the polynomials $G_{t}$ are $\delta$-vanishing on $Z$ and the algorithm terminates. Note that in this case the polynomials in $G^{t-1}$ may be no longer $\delta$-vanishing on $Z$ because $Z$ has been updated.
The next subsection introduces the reset framework, which resolves this situation. The way of reducing $\eta$ can affect the algorithm result. In
the present study, we decrease $\eta$ in a pragmatic way; we introduce a cooling parameter $\gamma<1$. Generally, $\eta$ was updated by $\gamma\eta$, but when the largest norm of the evaluation vector of $g\in G$, then $\eta$ was updated by that norm, \emph{i.e.}, $\eta=\min(\gamma\eta,\max_{g\in G}\|g(Z)\|)$.
The proper decrease of $\eta$ is left for future work. 

The iterative framework introduced in this section is summarized in Alg.~\ref{alg:ExactVanishPursuit}. In this subroutine, the orders of the data knotting and polynomial construction are reversed for easy implementation in the latter sections.

\begin{algorithm}[t]
\caption{ExactVanishPursuit\label{alg:ExactVanishPursuit}}
\begin{algorithmic}[1]
\Require{$G^{t}, F^{t}, \tilde{C}_t, Z, X_0, \varepsilon, \eta, \delta, \lambda$}
\Ensure{$G_t, F_t$}

\While{$\eta > \delta$ and $G_t$ is not empty}
 \State{Update $Z$ by solving Eq.~(\ref{eq:full-optimization-problem-pointwise});}
 \If{$\forall g \in G^t, \|g(Z)\| \le \delta$}
  \State{break}
 \EndIf
 \State{Decrease $\eta$;}
 \State{$G_t, F_t = \text{FindBasis}(\tilde{C}_t, F^{t-1}, Z, X_0, \varepsilon, \eta)$}
\EndWhile
\State{\Return $G_t,F_t$}
\end{algorithmic}
\end{algorithm}

\subsection{Algorithm\label{subsec:Algorithm}}

\begin{algorithm}[t]
\caption{Main\label{alg:EVAVI}}
\begin{algorithmic}[1]
\Require{
$X_0,\varepsilon, \delta, \lambda$
}
\Ensure{$G, Z$}
\State{\# Initializetoin}
\State{$G = \{\}, F = F_0 = \{f(\cdot) = 1/\sqrt{|X|}\}$}
\State{$\tilde{C}_1 = Z = X_0$}
\State{$\eta = \varepsilon, t=1$}
\Loop
 \State{\# Compute bases of degree-$t$ polynomials}
 \State{$G_t, F_t = \text{FindBasis}(\tilde{C}_t, F, Z, X_0, \varepsilon,\eta)$}
 \State{$G_t, F_t = \text{ExactVanishPursuit}(G\cup G_t, F\cup F_t, $}
 \State{\hspace{13.5mm}$\tilde{C}_t, Z, X_0, \varepsilon,\eta, \lambda)$}

 \State{$G = G \cup G_t, F = F \cup F_t$}
 \State{$C_t = \{f_t f_1; f_t\in F_t, f_1\in F_1\}$}
 \State{}

 \If{$C_t$ is empty}
  \If{$\forall g \in G, \|g(Z)\| \le \delta$}
   \State{\Return $Z, G$}
  \Else
   \State{\# Reset to degree 1}
   \State{$t=1$}
   \State{$G=\{\}, F = \{f(\cdot)=1/\sqrt{Z}\}$}     
   \State{$C_1 = Z$}     
   \State{Decrease $\eta$;}
  \EndIf   
 \Else 
  \State{$t = t+1$} 
 \EndIf
\EndLoop

\end{algorithmic}
\end{algorithm}

This section describes the overall algorithm of the proposed method (Alg.~\ref{alg:EVAVI}). The input are data points $X_{0}$, the error tolerances $\varepsilon,\delta$, and the regularization weight $\lambda$. The algorithm outputs a set of polynomials $G$ and data knots $Z$ for which polynomials in $G$ are $\varepsilon$-vanishing on $X_{0}$ and $\delta$-vanishing on $Z$. 

As it proceeds, the algorithm increments the degree of the polynomials. The degree-$0$ polynomial sets are initialized to
$G_{0}=\left\{ \right\}$, $F_{0}=\{f(\cdot)=1/\sqrt{|X|}\}$, and the initial data knots
$Z$ are set to $X_{0}$. We also introduce an error tolerance parameter $\eta$, which is set to $\eta=\varepsilon$. Although we aim to discover the $\delta$-vanishing polynomials on $Z$, we first consider the $\eta$-vanishing
polynomials on $Z$, which is updated rather than fixed. $\eta$ is gradually decreased throughout the iterations, and eventually reaches $\delta$, thereby obtaining $\delta$-vanishing polynomials.

At degree-$t$, the algorithm proceeds through the following four steps: (1) generate
$G_{t}$ and $F_{t}$ by Alg.~\ref{alg:FindBasis}, where $G_{t}$
is a set of polynomials that are $\varepsilon$-vanishing on $X_{0}$
and $\eta$-vanishing on $Z$; (2) update $G_{t}$, $F_{t}$, and
$Z$ by Alg.~\ref{alg:ExactVanishPursuit} such that the polynomials in
$G_{t}$ become $\delta$-vanishing on $Z$; (3) generate degree-$(t+1)$
candidate polynomials for the next iteration; (4) check the termination conditions
 (\emph{reset} or advance to the next degree). Reset restores all variables except $Z$ and $\eta$ to the $t=1$ stage. A reset is performed
if there is a $\delta$-nonvanishing polynomial in $G$ and
the algorithm cannot proceed to the next degree, \emph{i.e.}, when $C_{t}$
is empty. The reset system feedbacks the results of higher-degree polynomials
to lower-degree ones via the data knots $Z$. To our knowledge, the reset system is unique to our method; all of the existing methods appear to greedily construct the polynomials from lower to higher degrees.

Termination is guaranteed when $\eta$ reaches $\delta$. To prove it, first note that when $\eta=\delta$,
no data knotting occurs so $Z$ is fixed. To describe arbitrary polynomials on
$Z$, we need collect $|Z|$ linearly independent
polynomials in $F$, because the polynomials are associated with $\mathbb{R}^{|Z|}$-dimensional
vectors. In Alg.~\ref{alg:FindBasis}, the column space of the
evaluations of candidate polynomials $C_{t}$ is orthogonal to that
of $F$ on $Z$. By its construction, the column space of $F_{t}(Z)$
approximately spans the column space of $C_{t}(Z)$. When $\left|F_{t}\right|=0$,
the algorithm terminates; otherwise, the rank of $F$ is strictly
increased by appending $F_{t}$. Therefore, the rank of $F$ reaches
$\left|Z\right|$ after a finite number of steps, and the algorithm terminates.
As $\eta=\delta$, all polynomials in $G$ are $\delta$-vanishing
polynomials on $Z$.

Note that the output $G$ is not necessarily a basis of the vanishing ideal for $Z$ because polynomials that are $\delta$-vanishing on $Z$ but $\varepsilon$-nonvanishing on $X_{0}$ are excluded from $G$. This result is reasonable because the polynomials in $G$ do not well approximate the original data. In some cases, however, we require a basis of the vanishing ideal. Such a basis can be generated by applying existing basis generation methods such as VCA to small data knots $Z$, which is much less computationally costly than applying to $X_{0}$.

\section{Results}

In this sections, we demonstrate that our method discovers a compact
set of low-degree polynomials and a few data knots that well
represent the original points. The proposed method exhibits both noise tolerance and
good preservation of the algebraic structure.  We first illustrate the vanishing polynomials and data knots
obtained on simple datasets as qualitative evaluation. In the next classification task, we show that the polynomials output by out method avoid overfitting and hold the useful nonlinear feature of data as observed in VCA. 
Finally, we evaluate the representativeness of the data knots by training $k$-nearest neighbor classifiers in the classification tasks. 
Note that classification tasks are adopted to measure how well the proposed method can hold nonlinear structure of data. 
The proposed method is not specially tailored for classifications, and it can also contribute to other tasks where vanishing ideal based approach has been introduced.

\subsection{Illustration with simple data}

\begin{figure*}
\includegraphics[scale=0.27]{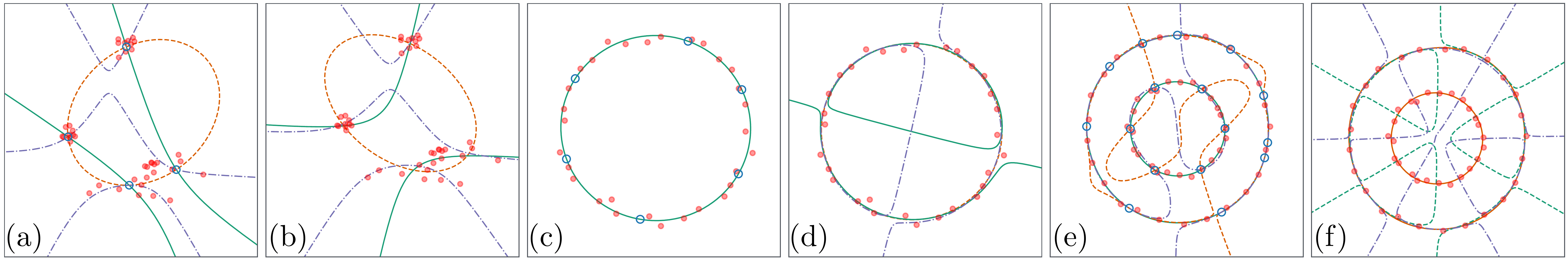}\caption{A set of vanishing polynomials and data knots (blue circles) output
by the proposed method (a,c,e) and VCA (b,d,f) for simple data exposed to noise (red dots).
The input data are (a,b) three blobs with different variance (c,d) a single circle, (e,f) two concentric circles. 
To enhance visibility, not all of the polynomials are shown.
\label{fig:result1}}
\end{figure*}


We applied our method and VCA with the same error tolerance to simple data perturbed by noise: three blobs
with different variances (60 points, 30\% noise on one blob;
5\% noise on the remaining blobs), a single circle
(30 samples, 5\% noise), and a pair of concentric circles (50 samples, 2\% noise). Here $n$\% noise denotes zero-mean Gaussian
noise with $n$-standard deviation. The blobs were generated by adding
noise at three distinct points. As shown in the Fig.~(\ref{fig:result1}),
each set of polynomials obtained by our method exactly vanishes on the data knots, and approximately
vanishes on the input data points, while those by VCA only coarsely intersect with each other. In Figs.~\ref{fig:result1}(a) and \ref{fig:result1}(b),
one blob has much larger variance than the others. Whereas each
blob with small variance is represented by a single data knot,
the very noisy blob is represented by two knots; consequently,
polynomials obtained by our method almost exactly vanish on the knots and approximately vanish over the whole blob. In contrast, while the polynomials obtained by VCA are similar to those by ours, they intersect with each other much more coarsely. In Figs.~\ref{fig:result1}(c), \ref{fig:result1}(d), \ref{fig:result1}(e), and \ref{fig:result1}(f), both our method and VCA discovered the lowest algebraic structures (a circle and a pair of concentric circles). In Fig.~\ref{fig:result1}(c), our method only outputs a circle since other polynomials that approximately vanish on the original data do not sufficiently vanish on the data knots. In Fig.~\ref{fig:result1}(e), some polynomials discovered by our method are different from those by VCA for better preserving the algebraic structure. Note that the same error tolerance is used for both our method and VCA. Thus, the polynomials obtained by our method still approximately vanish on the original points as those by VCA. 

\subsection{Compact lower-degree feature extraction}
\begin{table*}
\caption{Classification results\label{tab:Classification-results}}
\begin{tabular}{|c|c|c|c|c||c|c||c|c||c|c|}
\hline
 & \multicolumn{4}{c||}{Accuracy {[}\%{]}} 
 & \multicolumn{2}{c||}{Test runtime {[}sec{]}} 
 & \multicolumn{2}{c||}{\#features} 
 & \multicolumn{2}{c|}{\#degree}\tabularnewline
\hline
 & Proposed & VCA & Proposed-hd & VCA-hd & Proposed & VCA & Proposed & VCA & Proposed & VCA\tabularnewline
\hline
\hline
Iris & 0.96 & 0.95 & 0.75 & 0.59 & 3.6e-4 & 6.6e-4 & 14.9 & 62.8 & 1.5 & 2.0\tabularnewline
\hline
Wine & 0.98 & 0.98 & 0.94 & 0.67 & 7.6e-4 & 2.1e-3 & 100.5 & 592.9 & 1.7 & 2.3\tabularnewline
\hline
Vehicle & 0.80 & 0.80 & 0.53 & 0.60 & 1.6e-3 & 6.7e-2 & 121.5 & 4147.9 & 1.6 & 2.5 \tabularnewline
\hline
Vowel & 0.92 & 0.93 & 0.80 & 0.35 & 1.5e-3 & 2.3e-3 & 191.0 & 267.6 & 1.5 & 1.8\tabularnewline
\hline
\end{tabular}
\end{table*}



The vanishing polynomials obtained by our method compactly hold the original data
structure. To show this, we compare our method with VCA in four classification
tasks. Bothe methods were tested by Python implementation on a workstation with four processors and 8GB memory. Both the proposed method and VCA adopt the feature-extraction method of \citeauthor{livni2013vanishing}~(\citeyear{livni2013vanishing}):
\begin{align*}
\mathcal{F}(\mathbf{x}) & =\Bigl(\cdots,\underbrace{\left|g_{i}^{(1)}(\mathbf{x})\right|,\cdots,\left|g_{i}^{(|G_{i}|)}(\mathbf{x})\right|}_{G_{i}(\mathbf{x})^{\top}},\cdots\Bigr)^{\top},
\end{align*}
where $G_{i}=\{g_{i}^{(1)},...,g_{i}^{(|G_{i}|)}\}$ denote the computed
vanishing polynomials of the $i$-th class. By its construction, the feature $\mathcal{F}(\mathbf{x})$ of sample $\mathbf{x}$
in the $i$-th class should be a vector whose entries are approximately zero in the $G_{i}$ part and non-zero elsewhere. The classifier for the proposed method and VCA was a linear Support
Vector Machine~\cite{cortes1995support}. The datasets were downloaded
from~UCI machine learning repository~\cite{Lichman2013machine}. The hyperparameters were determined by
3-fold cross validation and the results were averaged over ten independent
runs. In each run, the datasets were randomly split into training (60\%) and
test (40\%) datasets.

The classification results are summarized in Table~\ref{tab:Classification-results}.
The proposed method achieved comparable classification
accuracy to VCA, with much more compact features. The dimensions of the feature vectors
obtained by our method were only 3--70\% those
of the VCA feature vectors. The degree of the discovered polynomials was also lower in the proposed method. Consequently, the test runtime was lower in our method than in VCA. This result
suggests that the proposed method well preserves the data structure even
after data knotting. As we insisted in the introduction, higher-degree polynomials can be sensitive to noisy data under traditional vanishing polynomial construction with fixed data points. To confirm this, we also evaluated the methods by restricting the polynomials for feature extraction to half of them in the higher-degree part (Proposed-hd and VCA-hd). As can be seen from Table~\ref{tab:Classification-results}, the higher-degree polynomials by our method lead to much higher classification accuracy than those by VCA, which suggests that our method provides noise-tolerant polynomials even in relatively higher degree while VCA does not. An exception is the result for the Vehicle dataset. In this case, our method provided greatly compact features (only 2\% dimension of those by VCA). Thus the 50 \% restriction at Proposed- may remove important features that contribute much to the accuracy. 

\subsection{Evaluating data knots}

\begin{table}

\caption{$k$-nearest neighbor classification\label{tab:Classification-results-knots}}
\begin{tabular}{|c|c|c|c||c|}
\hline 
 & \shortstack{Data\\ knots} & \shortstack{$k$-means\\ centroids} & \shortstack{Original\\ points} & \shortstack{Knotting\\ ratio}\tabularnewline
\hline 
\hline 
Iris & 0.95 & 0.95 & 0.94 & 0.08\tabularnewline
\hline 
Wine & 0.95 & 0.97 & 0.95 & 0.20\tabularnewline
\hline 
Vehicle & 0.60 & 0.61 & 0.68 & 0.05 \tabularnewline
\hline 
Vowel & 0.76 & 0.79 & 0.96 & 0.53\tabularnewline
\hline 
\end{tabular}

\end{table}

In the proposed framework, data knotting greatly reduces the number of original noisy
data points, enabling lower-degree
vanishing polynomials. Here we evaluate the representativeness of the data knots with the classification accuracy of the $k$-nearest neighbor classifier trained with data knots. As baselines, we also trained classifiers with $k$-means centroids and the original points as baselines. The number of centroids for $k$-means clustering is set to the same number of data knots for each class. As the number of data knots and $k$-means centroids are much smaller than the number of original points, we set $k=1$ in the classifiers. The other training and testing settings were those of the previous section.

The results are summarized in Table~\ref{tab:Classification-results-knots}.
The number ratio between the data knots and original points, called the knotting ratio, confirms the that the original data points were condensed into far fewer points after the knotting. Training the classifier on few data knots achieves the comparable accuracy of the classification with the $k$-means centroids, supporting our argument that the data knots well represent the original data. Note that the data knots are designed to provide lower-degree vanishing polynomials while $k$-means clustering is for simply summarizing nearby points. Compared to the classification results with original data points, those with data knots were comparable for the datasets with fewer classes (Iris and Wine; 3 classes). However, the accuracy degrades for the datasets with more classes (four classes in Vehicle and eleven classes in Vowel). In these datasets, there can be more overlapped points across different classes, resulting in accuracy degradation in ours and $k$-means. A possible solution specially tailored for classification is to introduce class-discriminative information to data knots. This modification is an interesting future extension.

\section{Conclusion and Future work}
The present paper focused on the tradeoff between noise-tolerance and better preserving an algebraic structure at the vanishing ideal construction, which has not been explicitly considered before. We addressed a new problem to discover a set of vanishing polynomials that approximately vanish
on the noisy input data and almost exactly vanish on the jointly discovered
representative data points (called data knots). In the proposed framework, we introduced a vanishing polynomial construction method that takes into account two different point sets with different noise-tolerance scales. We also linked the newly introduced nonlinear regularization term to the Mahalanobis distance, which is commonly used in metric learning. 
In experiments, the proposed method discovered much more compact and lower-degree algebraic systems than the existing method. 

Computing the vanishing polynomials that exactly vanish on the data knots remains an open problem. Exactly vanishing polynomials are desired because they can be manipulated by algebraic operations and combined with other algebraic tools. For practical reasons (numerical implementation and optimization), our method returns exactly vanishing polynomials in an extreme case only ($\delta=0$), which often show poor performance in reality due to the numerical instability. Another future work is to increase computational efficiency. The proposed method is rather slower than VCA in training runtime (see the supplemental material) due to the optimization step for data knotting. Also, our method has three main hyperparameters to be tuned ($\varepsilon$, $\delta$, and $\lambda$), which requires additional cost for the cross validation step. Empirically, the most important hyperparameter is $\varepsilon$ that defines the error tolerance for the original data. Determining $\varepsilon$ is similar to determining the number of principal components in Principal Component Analysis (PCA), while $\varepsilon$ affects not only selecting linear polynomials but also selecting nonlinear polynomials. How to choose $\varepsilon$ is still an open problem for many vanishing ideal based approaches including ours.

\section*{Acknowledgement}
This work was supported by JSPS KAKENHI Grant Number 17J07510. The authors would like to thank Hitoshi Iba, Hoan Tran Quoc, and Takahiro Horiba of the University of Tokyo for helpful conversations and fruitful comments.




\bibliographystyle{aaai}

\newpage


\section*{Supplementary material}

# Supplemental material for
## *Approximate Vanishing Ideal via Data Knotting*

## 1 Proof for the problem reformulation in Section 4.2

Here, we prove

$$C_t(Z) = \tilde{\mathbf{U}}\tilde{\mathbf{D}}\tilde{\mathbf{V}}^\top \iff F_t(Z) = F_t(X), G_t(Z) = \mathbf{O},$$

where we use the notations in the Section 4.2 in the main text.

We first show this relation from the left hand side to the right hand side. The following is Eq. (4) in the main text.

$$C_t(X) = \tilde{\mathbf{U}}\tilde{\mathbf{D}}\tilde{\mathbf{V}}^\top + \mathbf{U}_\varepsilon \mathbf{D}_\varepsilon \mathbf{V}_\varepsilon^\top,$$

where $\tilde{\mathbf{V}}^\top \tilde{\mathbf{V}} = \mathbf{I}$, $\mathbf{V}_\varepsilon^\top \mathbf{V}_\varepsilon = \mathbf{I}$, and $\mathbf{V}_\varepsilon^\top \tilde{\mathbf{V}} = \mathbf{O}$. Multiplying both sides of $C_t(Z) = \tilde{\mathbf{U}}\tilde{\mathbf{D}}\tilde{\mathbf{V}}^\top$ by $\tilde{\mathbf{V}}$ and $\mathbf{V}_\varepsilon$ respectively, we obtain

$$C_t(Z)\tilde{\mathbf{V}} = \tilde{\mathbf{U}}\tilde{\mathbf{D}} = C_t(X)\tilde{\mathbf{V}},$$
$$C_t(Z)\mathbf{V}_\varepsilon = \tilde{\mathbf{U}}\tilde{\mathbf{D}}\tilde{\mathbf{V}}^\top \mathbf{V}_\varepsilon = \mathbf{O}.$$

For the first equation, recall that $F_t$ is rescaled by the evaluation on $Z$ (let the scaling factor be $\mathbf{S}$). Then, $F_t(Z) = C_t(Z)\tilde{\mathbf{V}}\mathbf{S} = C_t(X)\tilde{\mathbf{V}}\mathbf{S} = F_t(X)$. For the second equation, by its definition, $G_t(Z) = C_t(Z)\mathbf{V}_\varepsilon = \mathbf{O}$.

Next, we show the relation to prove from the right hand side to the left hand side. From $F_t(Z) = F_t(X), G_t(Z) = \mathbf{O}$,

$$C_t(Z)\tilde{\mathbf{V}} = C_t(X)\tilde{\mathbf{V}},$$
$$C_t(Z)\mathbf{V}_\varepsilon = \mathbf{O}.$$

By horizontally stacking above equations,

$$C_t(Z)\begin{pmatrix}\tilde{\mathbf{V}} & \mathbf{V}_\varepsilon\end{pmatrix} = \begin{pmatrix}C_t(X)\tilde{\mathbf{V}} & \mathbf{O}\end{pmatrix}.$$

Since $(\tilde{\mathbf{V}} \ \mathbf{V}_\varepsilon)$ is a square matrix, $(\tilde{\mathbf{V}} \ \mathbf{V}_\varepsilon)(\tilde{\mathbf{V}} \ \mathbf{V}_\varepsilon)^\top = \mathbf{I}$. Therefore,

$$\begin{aligned}
C_t(Z) &= \left(C_t(X)\tilde{\mathbf{V}} \ \mathbf{O}\right) \begin{pmatrix} \tilde{\mathbf{V}} & \mathbf{V}_\varepsilon \end{pmatrix}^\top, \\
&= C_t(X)\tilde{\mathbf{V}}\tilde{\mathbf{V}}^\top, \\
&= \left(\tilde{\mathbf{U}}\tilde{\mathbf{D}}\tilde{\mathbf{V}}^\top + \mathbf{U}_\varepsilon \mathbf{D}_\varepsilon \mathbf{V}_\varepsilon^\top\right)\tilde{\mathbf{V}}\tilde{\mathbf{V}}^\top, \\
&= \tilde{\mathbf{U}}\tilde{\mathbf{D}}\tilde{\mathbf{V}}^\top\tilde{\mathbf{V}}\tilde{\mathbf{V}}^\top + \mathbf{U}_\varepsilon \mathbf{D}_\varepsilon \mathbf{V}_\varepsilon^\top \tilde{\mathbf{V}}\tilde{\mathbf{V}}^\top, \\
&= \tilde{\mathbf{U}}\tilde{\mathbf{D}}\tilde{\mathbf{V}}^\top.
\end{aligned}$$

## 2 The detail derivation of the generalized Mahalanobis distance

We show the detail derivation of the Mahalanobis distance in $t = 1$ case that was omitted in the main text. The same derivation holds in $t > 1$ cases, so we keep using $t$ for general discussion. Again, the notations are the same as those in the main text. At degree $t$ the data $X$ is mapped to $C_t(X)$. For $t = 1$, $C_1(X)$ is a mean-centralized $X$. The covariance matrix $C_t(X)^\top C_t(X)$ is approximated by its principal variance $\Sigma$, which consists of singular vectors corresponding to singular values exceeding $\delta$.

$$\begin{aligned}
\Sigma &= \left(\hat{\mathbf{U}}\tilde{\mathbf{E}}\tilde{\mathbf{W}}\tilde{\mathbf{V}}_0 \ \ \mathbf{U}\tilde{\mathbf{D}}\tilde{\mathbf{V}}^\top \mathbf{V}_0^{\varepsilon\top}\right)^\top \left(\hat{\mathbf{U}}\tilde{\mathbf{E}}\tilde{\mathbf{W}}\tilde{\mathbf{V}}_0 \ \ \mathbf{U}\tilde{\mathbf{D}}\tilde{\mathbf{V}}^\top \mathbf{V}_0^{\varepsilon\top}\right), \\
&= \tilde{\mathbf{V}}_0^\top \tilde{\mathbf{W}}\tilde{\mathbf{E}}\hat{\mathbf{U}}^\top \hat{\mathbf{U}}\tilde{\mathbf{E}}\tilde{\mathbf{W}}\tilde{\mathbf{V}}_0 + \mathbf{V}_0^{\varepsilon\top}\tilde{\mathbf{V}}\tilde{\mathbf{D}}\mathbf{U}^\top \mathbf{U}\tilde{\mathbf{D}}\tilde{\mathbf{V}}\mathbf{V}_0^\varepsilon, \\
&= \tilde{\mathbf{V}}_0^\top \tilde{\mathbf{W}}\tilde{\mathbf{E}}^\top \tilde{\mathbf{E}}\tilde{\mathbf{W}}\tilde{\mathbf{V}}_0 + \mathbf{V}_0^{\varepsilon\top}\tilde{\mathbf{V}}\tilde{\mathbf{D}}^\top \tilde{\mathbf{D}}\tilde{\mathbf{V}}\mathbf{V}_0^\varepsilon.
\end{aligned}$$

Using $\tilde{\mathbf{V}}_0^\top \tilde{\mathbf{V}}_0 = \mathbf{I}, \tilde{\mathbf{W}}^\top \tilde{\mathbf{W}} = \mathbf{I}, \mathbf{V}_0^{\varepsilon\top}\mathbf{V}_0^\varepsilon = \mathbf{I}$, and $\tilde{\mathbf{V}}^\top \tilde{\mathbf{V}}$, we obtain

$$\left(C_t(X)^\top C_t(X)\right)^{-1} \approx \tilde{\mathbf{V}}_0^\top \tilde{\mathbf{W}}(\tilde{\mathbf{E}}^\top \tilde{\mathbf{E}})^{-1}\tilde{\mathbf{W}}\tilde{\mathbf{V}}_0 + \mathbf{V}_0^{\varepsilon\top}\tilde{\mathbf{V}}(\tilde{\mathbf{D}}^\top \tilde{\mathbf{D}})^{-1}\tilde{\mathbf{V}}\mathbf{V}_0^\varepsilon.$$

This leads to the result of $\|F_1(\mathbf{x}) - F_1(\mathbf{y})\|^2 = (\mathbf{x} - \mathbf{y})^\top \Sigma^{-1}(\mathbf{x} - \mathbf{y})$ in the main text. For the nonlinear case, the $(\mathbf{x} - \mathbf{y})$ part becomes $(C_t(\mathbf{x}) - C_t(\mathbf{y}))$.

## 3 Training runtime

As mentioned in Section 6 of the main text, the proposed method takes much more time to train than VCA. This is because our method requires solving optimization problems at data knotting stage, which plays an important role in our method to pursue the algebraic soundness and compact polynomial set. The training time in the classification tasks in Section 5.2 is shown in Table S1. The accuracy and test runtime are those shown in Table 1 in the main text. Note that the main purpose of this experiment was to show that the compact set of polynomials obtained by the proposed methods holds as good nonlinear structures of data as those by VCA. Compactness can be more important than

Table S1: Classification results

|  | Accuracy [%] |  | Training runtime [sec] |  | Test runtime [sec] |  |
| --- | --- | --- | --- | --- | --- | --- |
|  | Proposed | VCA | Proposed | VCA | Proposed | VCA |
| Iris | 0.96 | 0.95 | 5.3 | 1.4e-2 | 3.6e-4 | 6.6e-4 |
| Wine | 0.98 | 0.98 | 60 | 1.4e-1 | 7.6e-4 | 2.1e-3 |
| Vehicle | 0.80 | 0.80 | 18e+1 | 5.4 | 1.6e-3 | 6.7e-2 |
| Vowel | 0.92 | 0.93 | 30 | 3.6e-2 | 1.5e-3 | 2.3e-3 |

runtime when the goal of the task is to discover intrinsic nonlinear relation between variables (e.g., polynomial dynamical system identification [Kera and Hasegawa (2016)] and nonlinear data modeling in robotics [Iraji and Chitsaz (2017)]). If one tries to tailor our method for the classification task, runtime improvement and accuracy improvement are desired. In this case, we can reduce the number of data by preprocessing such as PCA and matrix sketching (e.g., Liberty (2013)). Accuracy improvement can be achieved by introducing discriminative information as (Hou, Nie, and Tao (2016)).



\end{document}